\documentclass[conference]{IEEEtran}
\IEEEoverridecommandlockouts
\usepackage{cite}
\usepackage{amsmath,amssymb,amsfonts}
\usepackage{algorithmic}
\usepackage{graphicx}
\usepackage{textcomp}
\usepackage{xcolor}
\usepackage[colorlinks,linkcolor=blue]{hyperref}
\begin{document}

\title{Egocentric Hand-object Interaction Detection\\

\thanks{This work was supported by National Natural Science Foundation of China (Grant No. 62202285). \\
*Corresponding author.
}

}

\author{\IEEEauthorblockN{ Yao Lu}
\IEEEauthorblockA{\textit{University of Bristol, Bristol, UK} \\
\textit{Phillips Research, Shanghai, China}\\
sunny.lu@philips.com}
\and
\IEEEauthorblockN{Yanan Liu\textsuperscript{*}}
\IEEEauthorblockA{\textit{School of Microelectronics, Shanghai University, Shanghai, China}\\
\textit{Bristol Robotics Laboratory, University of Bristol, Bristol, UK} \\
yanan.liu@ieee.org}
}

\maketitle

\begin{abstract}
In this paper, we propose a method to jointly determine the status of hand-object interaction. This is crucial for egocentric human activity understanding and interaction. From a computer vision perspective, we believe that determining whether a hand is interacting with an object depends on whether there is an interactive hand pose and whether the hand is touching the object. Thus, we extract the hand pose, hand-object masks to jointly determine the interaction status. In order to solve the problem of hand pose estimation due to in-hand object occlusion, we use a multi-cam system to capture hand pose data from multiple perspectives. We evaluate and compare our method with the most recent work from Shan et al. \cite{Shan20} on selected images from EPIC-KITCHENS \cite{damen2018scaling} dataset and achieve $89\%$ accuracy on HOI (hand-object interaction) detection which is comparative to Shan's ($92\%$). However, for real-time performance, our method can run over $\textbf{30}$ FPS which is much more efficient than Shan's ($\textbf{1}\sim\textbf{2}$ FPS). A demo can be found from \url{https://www.youtube.com/watch?v=XVj3zBuynmQ}.
\end{abstract}


\section{Introduction}
\label{sec:intro}
Detecting interaction is attractive as a way to remove redundant information in a video sequence. In egocentric perception, extracting the HOI (hand-object interaction) can be crucial in action localisation and understanding in ADLs (Activities of Daily Living) or industrial applications. This competence is also useful in the context of Augmented Reality (AR) and Mixed Reality (MR), as a number of head-mounted devices generate vast volumes of egocentric footage that must be processed in order to be legitimate. A step in this approach is automating information extraction from egocentric video in the form of graphical collections or key-frames. In terms of content, key-frame (representative frame) extraction entails extracting the most informative frames that encapsulate the essential events in a video \cite{lei2014novel}. For storage, indexing, and retrieval, it is necessary to remove duplicate information from long video sequences \cite{rani2020social}. More importantly, detecting HOI provides a new way of thinking about video content understanding and content authoring, which have been a bottleneck in AR content production. Traditional AR or MR content is created through a rigorous process of model building, rigorous 3D registration and process design, which has the advantage of giving the user the most immersive effect possible, but also has the disadvantage of being cumbersome, time-consuming and unfriendly to non-specialists. With HOI detection, the content of the video can be automatically broken down and extracted, especially key steps that can be segmented and re-presented to the user. The whole process can be fully unsupervised. Although the content is not as polished as specially produced content, it allows content authoring to be automated and applied on a large scale.

\begin{figure}[t]
\centering
\includegraphics[width=1\columnwidth]{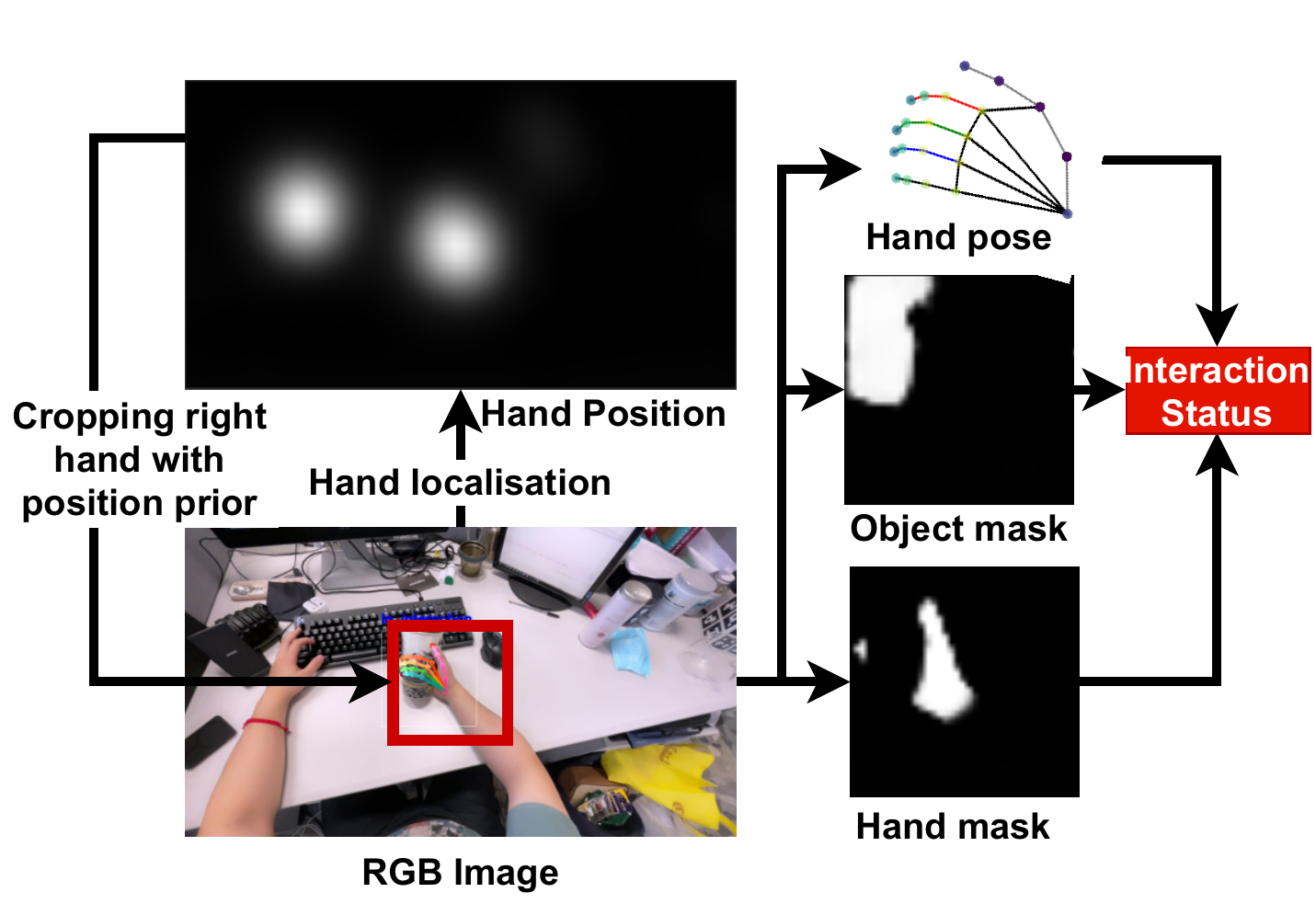} 
\caption{The pipeline of full hand-object interaction detection system.}
\label{fig:HOI_system}
\end{figure}

It is important to note that accurate estimation of hand pose is also an important factor in achieving high precision hand-object interaction. However, the egocentric vision also brings many challenges to hand pose estimation. Like hand-hand, hand-object and self-occlusion could bring huge uncertainty on hand pose estimation. Furthermore, to our knowledge, besides the work of Garcia et al. \cite{garcia2018first}, there are no real dataset targeting handling hand-object interaction problems. In this work, we collected a dataset containing hand pose ground truth under hand-object interaction using a method based on a multi-cam system and a clever pair-wise style for occlusion removal.

The main contribution of this paper focuses on two aspects. The first is the solution to the problem of hand pose estimation due to hand occlusion in egocentric vision. The second is the ability to quickly and accurately detect hand-object interaction by collecting relevant cues. In Section \ref{sec:related}, we review the works related to the both. In Section \ref{sec:hoi}, we detail the hand pose estimation module and HOI detection module respectively.

\section{Related Work}
\label{sec:related}
\subsection{HOI Detection}
Detecting the hand-object interaction is important for an egocentric based application. In Shan et al \cite{Shan20,gkioxari2018detecting}, the task of hand-object interaction detection is treated as object detection, different hand related cues are treated as different object classes, and whether an object is being interacted with or not is also treated as different object classes in training. In addition to this, in the work from Chen et al. \cite{chen2017}, with the depth sensor, they compare the depth within the palm area and the depth around the palm area to determine weather the hand is touching an object. Similarly, Lu et al. \cite{lu2019higs} use the same strategy and they plot an AUC (area under curve) curve to determine the best threshold for the criterion of 'touch' detection. Likitlersuang et al. \cite{likitlersuang2016interaction} put attention on higher level features, they build a hand-object interaction classifier with the optical flow and hand shape as input. The work from Schroder et al. \cite{schroder2017hand} use hand and object mask to predict the hand-object interaction status. The idea of using masks inspires our work; Differently, we explore the relationship between the hand-object pair and interaction with hand pose and masks. We also design our approach to be generic in that the method does not explicitly require a prior description of what objects are involved in an application to be tested. 

\subsection{Hand Pose Estimation under HOI}
To solve the 3-d hand pose estimation problem in egocentric perspective, people annotated real data \cite{rogez2015understanding,mueller2017real,garcia2018first} and created synthetic data \cite{mueller2017real,mueller2018ganerated,lin2021two,malik2019simple}. All of them provide 3-d hand pose ground truth. However, the type and quality of annotations vary. From the summary of table \ref{tab:dataset}, we found the real data an egocentric perspective is very scarce. The UCI-EGO \cite{rogez2015understanding} is annotated by manual refinement. FHAD \cite{garcia2018first} provides massive labelled 3-d hand-object manipulation data. However, the RGB frames are irreparably damaged by the magnetic sensors attached on hand for data collection. And for EgoDexter \cite{mueller2017real}. The quantity and quality of data annotated are very limited. For each frame, only the visible fingertips are manually labelled. Because the annotator labelled from depth images, the dis-alignment between the depth and RGB also leads to inaccuracies. The synthetic datasets provide abundant poses, objects and backgrounds, and they still suffer from the unrealistic shapes, poses and skin textures. Providing high-quality hand joints annotations in an egocentric perspective still remains challenging. 

\begin{table}
\begin{center}
\begin{tabular}{|c|c|c|c|c|c|}
\hline
 Dataset & S/R  & HOI   & Frames \\
\hline\hline
UCI-EGO\cite{rogez2015understanding}          & Real & None   & 400         \\
SynthHands\cite{mueller2017real}       & Synth & both   & 63,530   \\
EgoDexter\cite{mueller2017real}       & Real & obj   & 1485     \\
GANerated Hands\cite{mueller2018ganerated}     & Synth & both   & 330k    \\
FHAD\cite{garcia2018first}            & Real & obj   & 100k     \\
SynHandEgo\cite{malik2019simple}      & Synth & None   & -         \\
Ego3DHands\cite{lin2021two}     & Synth & None   & 50k/5k      \\

\hline
\end{tabular}
\end{center}
\caption{The table shows the recent datasets with egocentric data for hand pose estimation. 'S/R': Synthetic or Real dataset. 'HOI': Contains hand-object interaction data.}
\label{tab:dataset}
\end{table}

\section{Hand Interaction Detection System}
\label{sec:hoi}
We solve the hand-object interaction detection by utilising the contextual information of hand and object. More specifically, we use the hand pose, the hand mask and the object mask as cues to determine the interaction status. It can be expressed as:
$$P_{hoi} = f(H(pose), H(hand), H(object)) $$where $P_{hoi}$ is the possibility of hand-object interaction, and $H$ refers to the heatmap. The following content of the section details the implementation of the sub-tasks and how they are assembled as a system.

\subsection{Hand Pose Estimation}
We aim to predict the 3-d hand pose in an egocentric view, especially for the cases of hand-object interaction. To achieve that, we contribute the work in two steps: data collection and training. For data collection, we setup a multi-cam system that utilising multiple monocular cameras and 2-d keypoints detectors to acquiring 3-d hand pose annotation with constrains and optimisation. These collected 3-d data are re-projected back to each camera view and used for refining the 2-d hand keypoints detectors. The iterative pipeline was first used in \cite{simon2017hand} and improved in \cite{zimmermann2019freihand, hampali2020honnotate}. After obtaining the optimised 2-d hand detectors, we use a novel method that can significantly reduce the uncertainties caused by object occlusion. The data is collected in a pair-wise style.

\subsubsection{Data Capturing}
\label{sec:pose_data}
To overcome the data shortage on HOI, we set up a multi-cam system for data capturing. Different from other multi-cam systems like \cite{simon2017hand,zimmermann2019freihand}, we use the ArUco \cite{romero2018speeded} cube for online calibration. The cube returns the 6d pose of the camera relative to the cube centre. This enables us to capture from any view without frequent system re-calibration. Notably, our system can mimic the egocentric vision for different scenarios (head-based egocentric, chest-based egocentric or shoulder-based egocentric). 

\begin{figure}[t]
\centering
\includegraphics[width=1\columnwidth]{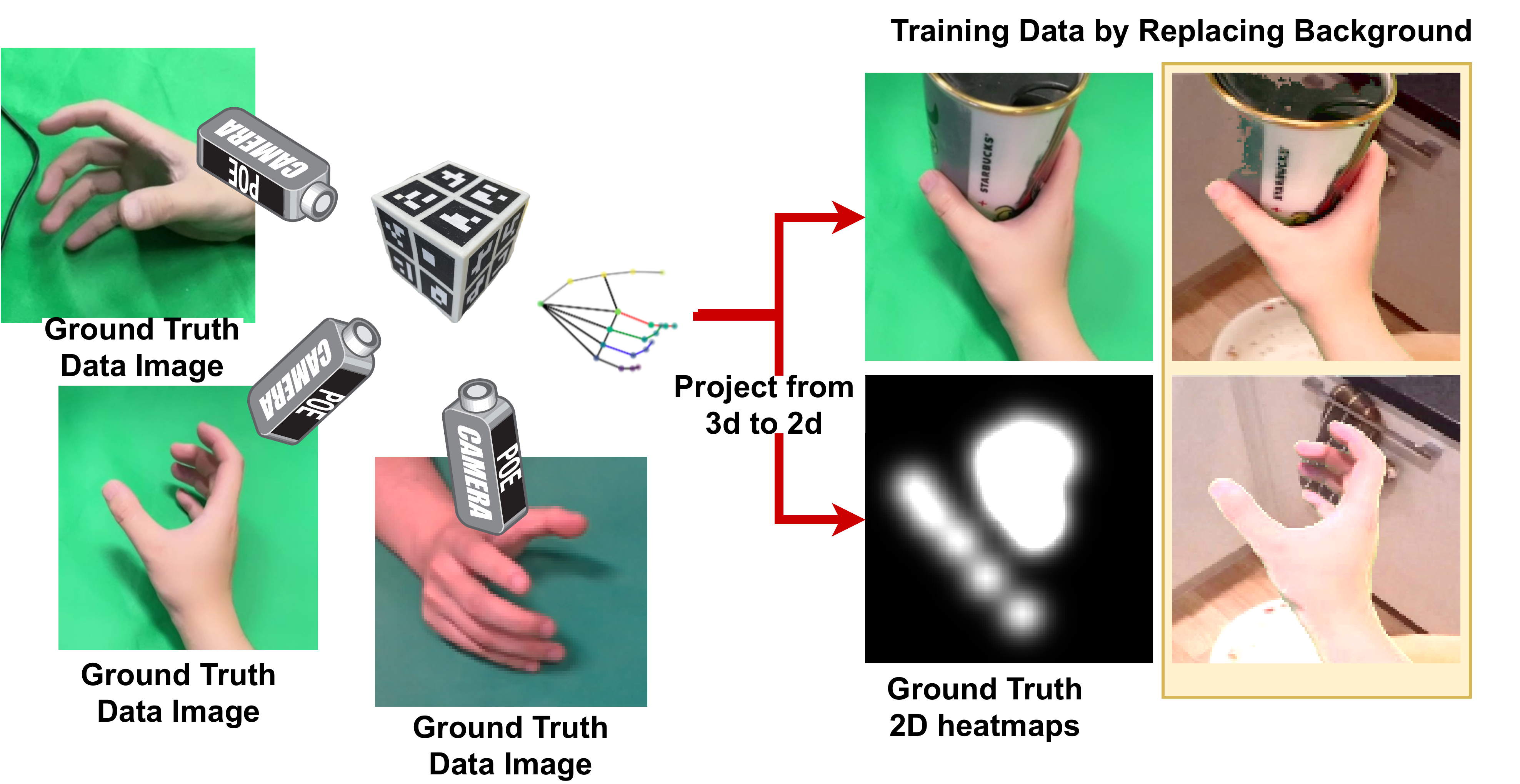} 
\caption{Data annotation process. First, we capture the object-in-hand poses as the target data to label. Then, we remove all the objects and keep the hand still to capture the source of ground truth. With a green screen, we can easily replace the background with other images.}
\label{fig:data_displaying}
\end{figure}

In our setup, we use $\textbf{3}$ cameras in our system. Apparently, $\textbf{3}$ is not enough to eliminate the uncertainties caused by object occlusion. However, it is easier for the hands without objects to identify the location of joints with a well-trained 2-d pose detector. By gathering all the detected 2-d poses from different views, the 3-d pose can be easily recovered by minimising the discrepancies observed by different cameras. Thus, we capture the hands with objects as data images, and then we remove the object while keeping hands as still as possible for ground truth data image capturing. The process is shown in figure \ref{fig:data_displaying}.

\subsubsection{Data Processing and Training}
We obtain two sets of images after the multi-cam system capturing from the previous step. Set $1$ has the images with the object in hand, and the set $2$ has the same hand pose with the object removed. For each image from set $2$, we sent it to a hand detector proposed by \cite{xiao2018simple}, and training on the datasets from the synthetic dataset GANerated Hand \cite{mueller2018ganerated}. Each view produces a set of heatmaps indicating the 2-d hand joint positions $j^{2d}$. For each joint, we have a corresponding pre-posed point $j^{3d}$ in 3-d space, its projections $j^{3d_proj}$ on all image planes should have minimum sum distance with the corresponding detected 2-d joints $j^{2d}$. The loss can be described as:
$$Loss_{proj} =\omega_{v}^{i}\sum_{k}\sum_{i}||j^{2d}_{i}-j^{3d\_proj}_i||_2$$where $\omega_{v}^{i}$ is the $i_{th}$ joint's confidence from the 2-d joint detector of the view $v$. We use Levenberg–Marquardt algorithm to optimise the loss function. If the loss can be smaller than a threshold, we consider the data valid and take the optimised 3-d joints as ground truth. The 2-d joints of set $1$ (with object) can be obtained by projecting the 3-d joints to the 2-d plane.

\subsubsection{Ego-Siam Dataset and Data Augmentation}
\label{sec:data_aug}
\begin{figure}
  \includegraphics[width=1\linewidth, height=0.5\linewidth]{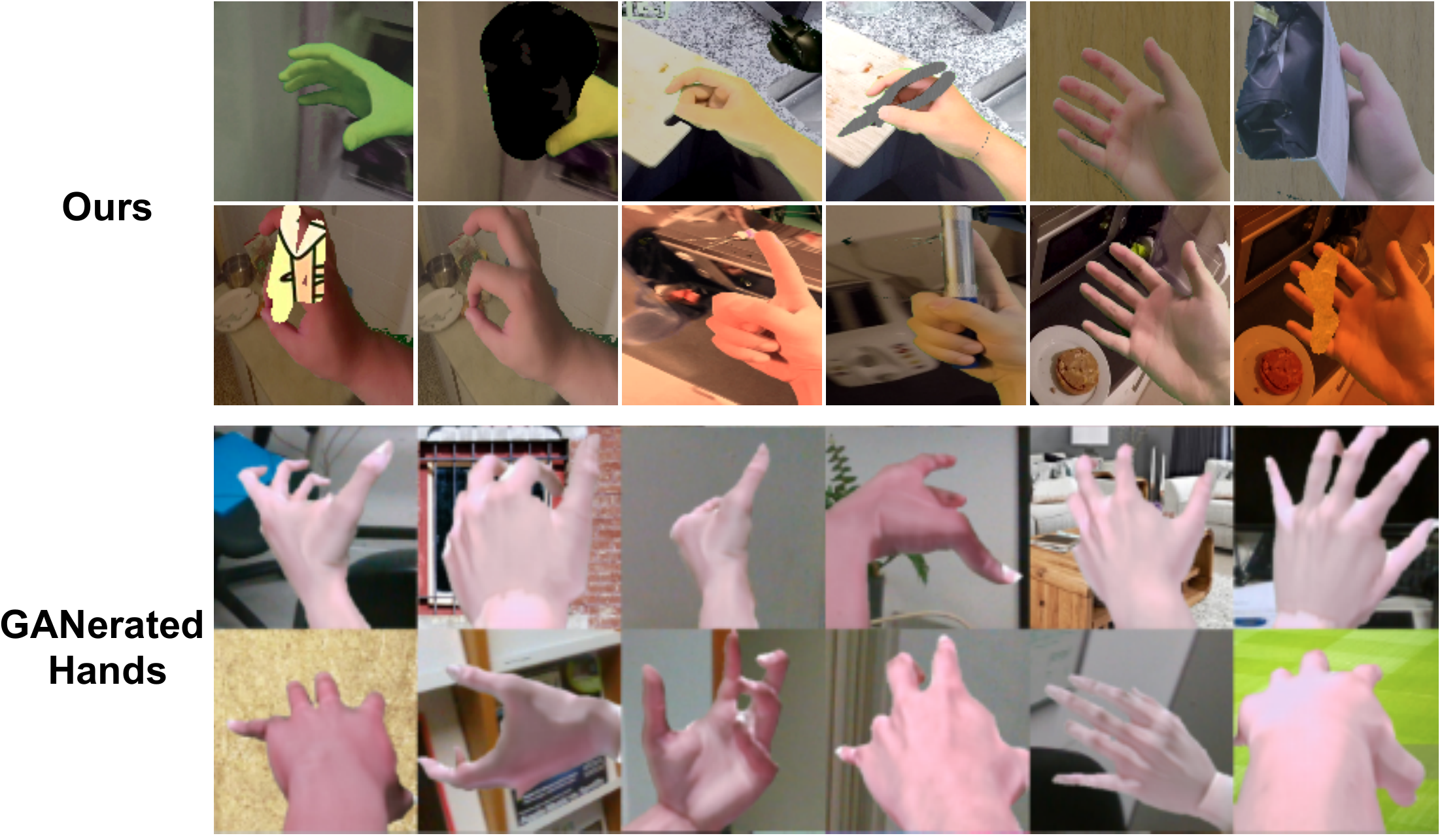}
  \caption{Visual comparison between our augmented data and the GANerated Hands \cite{mueller2018ganerated}. Our dataset is more realistic and involves more hand-object interactions.}
  \label{fig:data_aug}
\end{figure}

We collect $2k$ pairs of hand-object interaction data with a single right hand from a male. To gain more diversity on hand shapes and colours, another $2k$ frames without hand-object interaction are collected with $2$ male hands and $2$ female hands. There are $6k$ frames in total as training data. For testing, we collect another $500$ frames with hands performing different grasp types (the hand did not appear in the training set). With the green screen, the data augmentation can introduce more variability in the background. We identify the green colour in HSV colour space and replace it with real egocentric scenes, including the scenes from EPIC-KITCHENS \cite{damen2018scaling}, GTEA \cite{fathi2011learning} and the frames we collected from our offices and kitchens. Some backgrounds in skin colour are also added. Besides, we manually label the object masks for several hundreds of frames and replace them to increase the variability on object appearance. Another important augmentation is putting artificial occlusion on images. We randomly add line linkages between joints and circles with random sizes to simulate object occlusion. Other standard operations like random contrast, random brightness and random warp are also applied. All the augmentations and texture replacements run with the process of training. A comparison between our augmented data and the GANerated Hand \cite{mueller2018ganerated} is shown in figure \ref{fig:data_aug}. we have more realistic visual effects and extra artificial occlusions.

\subsection{In-hand Object Segmentation}
Segmenting hand and object are two topics that have been well explored separately. However, in-hand object segmentation is a complex and mostly overlooked problem of relevance in HOI (hand-object interaction) research. The main challenge could be the lack of training data and explicit applications as motivation. However, we argue, in-hand object segmentation is a strong cue for hand-object interaction analysis and can help with reducing the difficulty of in-hand object recognition for MR applications. Moreover, the contextual relationship between hand and object can be automatically encoded into the network by learning from the real hand-object interactions. As for hand-object interaction detection, this can be very helpful with identifying the non-interaction contact between hand and object.

\subsubsection{Dataset for Training and Evaluation}
\label{sec:data_anno}
\begin{figure}
\centering
\includegraphics[width=0.9\columnwidth]{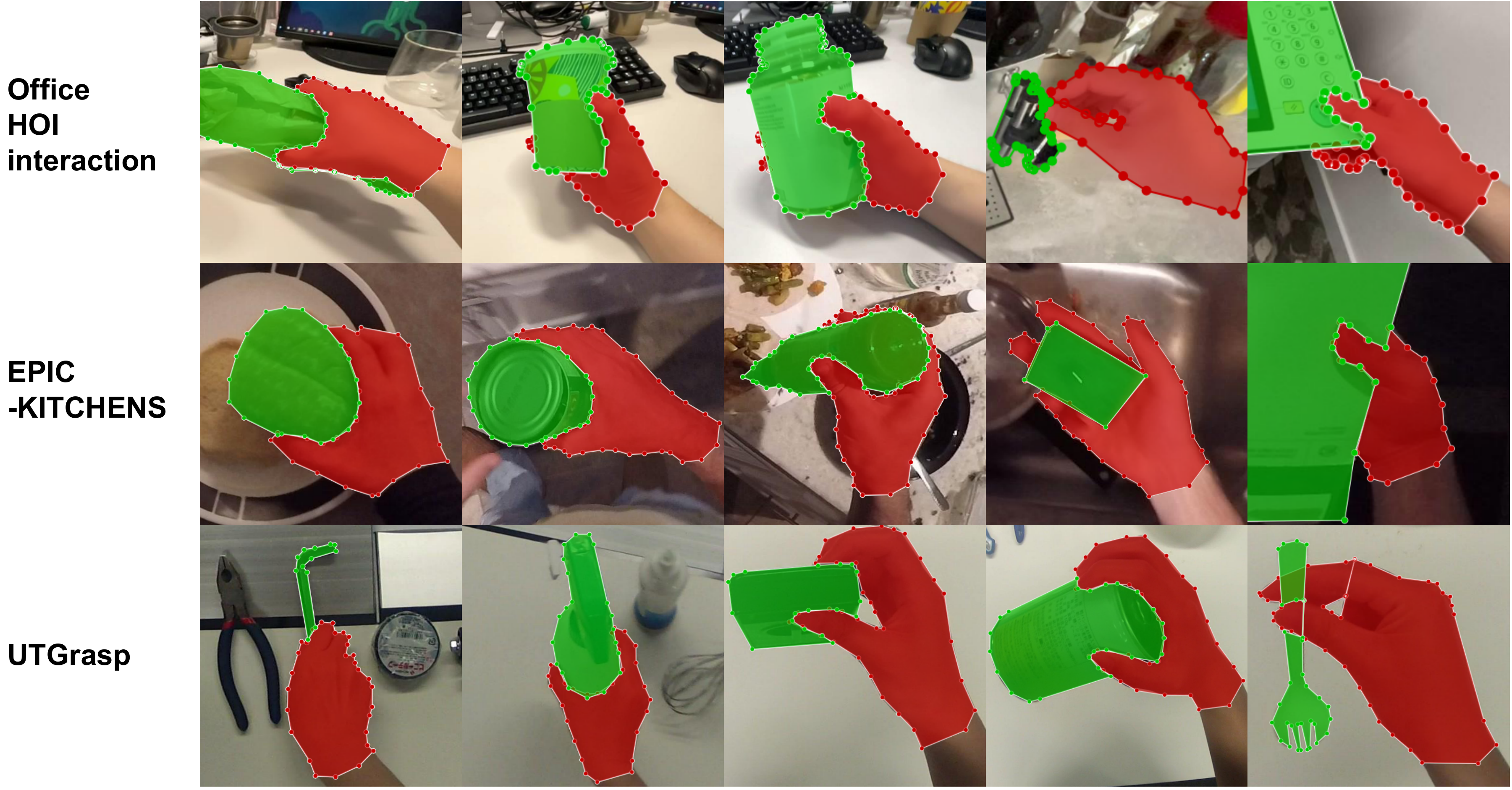} 
\caption{Examples of annotations from different datasets. The top row is our self-collected office hand-object interactions. The middle row is from EPIC-KITCHENS \cite{damen2018scaling}, and the bottom row comes from UTGrasp dataset \cite{cai2015scalable}.}
\label{fig:annotation}
\end{figure}

In order to obtain realistic data on hand-object interaction, our data consists of two parts. The first part comes from the previously annotated $6k$ images (\textbf{Ego-Siam}, detailed in section \ref{sec:data_aug}) for hand pose estimation. Another part of the training data is our new labelled dataset \textbf{GraspSeg}. The raw images in \textbf{Ego-Siam} are captured with the green background, we obtain the coarse segmented in-hand object mask by subtracting the green colour and skin colour from the raw images. While the new collected \textbf{GraspSeg} provides more refined training samples. We manually annotated the \textbf{GraspSeg} containing $3k$ images from various sources, including EPIC-KITCHENS \cite{damen2018scaling}, UTGrasp \cite{cai2015scalable} and our self-captured data in office and kitchen environments. We intentionally chose images that contain different types of hand-object interactions to broaden the data variety. More importantly, we also choose the hands without hand-object interaction or non-interaction contacting as negative samples. All the data images are annotated with the hand mask and in-hand object mask. Examples are shown in figure \ref{fig:annotation}.


\begin{figure}
\centering
\includegraphics[width=1\columnwidth]{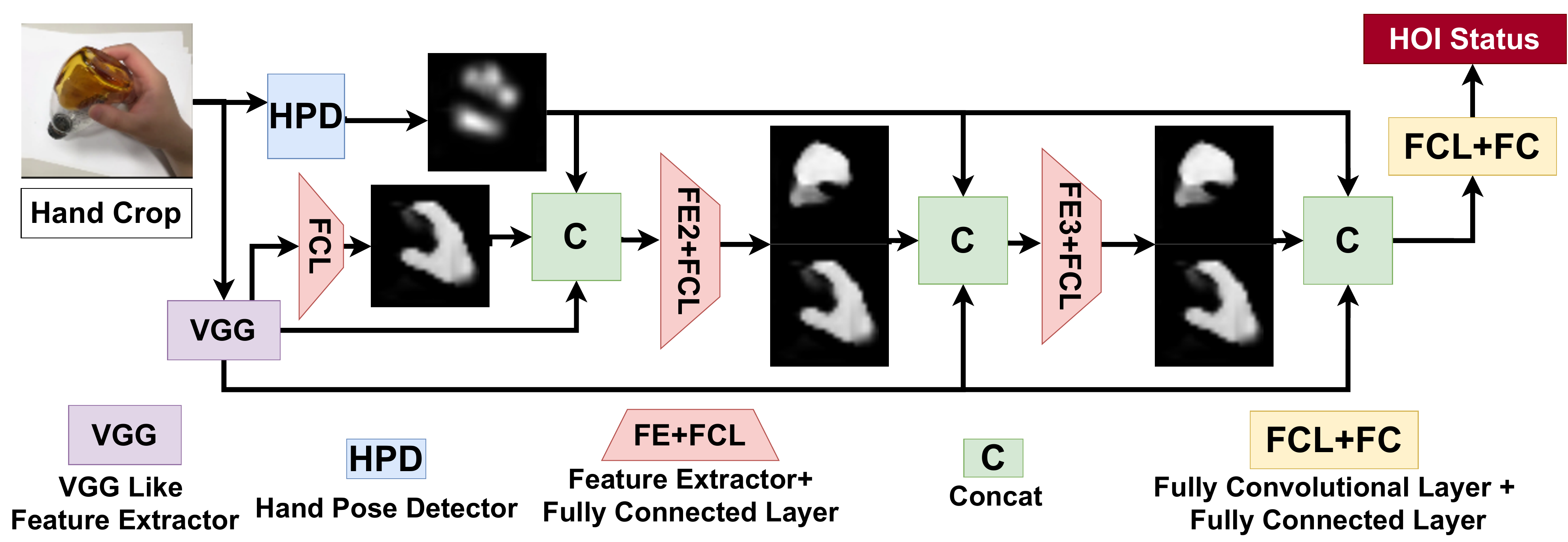} 
\caption{The schematic of our network. We use a 3-layer cascade structure to feed hand pose cues and hand mask cues to predict the 'Grasp Response Map (object mask)' and the interaction status. }
\label{fig:GRM}
\end{figure}

\subsubsection{Grasp Response Map and Interaction Detection}
Instead of segmenting the in-hand object mask directly, we propose a new attribute that not only predict the object area, it also implicitly reflects the interaction status, we call the attribute GRM (Grasp Response Map) and it is generated by our network which is shown in figure \ref{fig:GRM}. It is cascaded structured and contains three stages. The first stage predicts the hand mask with FCL (fully convolutional layer) \cite{long2015fully}. The input for the second stage is concatenated by the predicted hand mask, hand pose heatmaps and extracted features from the backbone. The object masks $m_{obj}$ is produced together with the hand mask from the stage $2$ and $3$. It can be regarded as a possibility distribution with the joint conditions of hand pose $H_{k(p)}$, hand mask $M_{hand}$ and the corresponding visual features. We call the predicted object mask $M_{obj}$ GRM (Grasp Response Map). At the end of the network, the hand cues and object cues jointly determine the hand-object interaction status. In our network design, we add a small, fully convolutional layer to reduce the dimension of the features and predict the possibility of hand-object interaction with a fully connected layer head. 

\begin{table}
\begin{center}
\begin{tabular}{|l|c|c|c|c|c}
\hline
  & S Clear & U Clear & S Cluttered & U Cluttered \\
\hline\hline
IOU & \textbf{0.84} & 0.77& 0.78& 0.64 \\
PA & \textbf{0.86} & 0.79& 0.85& 0.74 \\
\hline
\end{tabular}
\end{center}
\caption{The table shows the quantitative results of object segmentation under different conditions. 'S' stands for 'seen' and 'U' stands for 'unseen'.}
\label{tab:rst}
\end{table}

\section{Experiments and Results}

In this section, we evaluate the performance of three main components, namely hand pose estimation, in-hand object segmentation and hand-object interaction detection. For the test dataset, we used some publicly available datasets as well as the test set we collected.

\begin{figure}[t]
  \includegraphics[width=\linewidth]{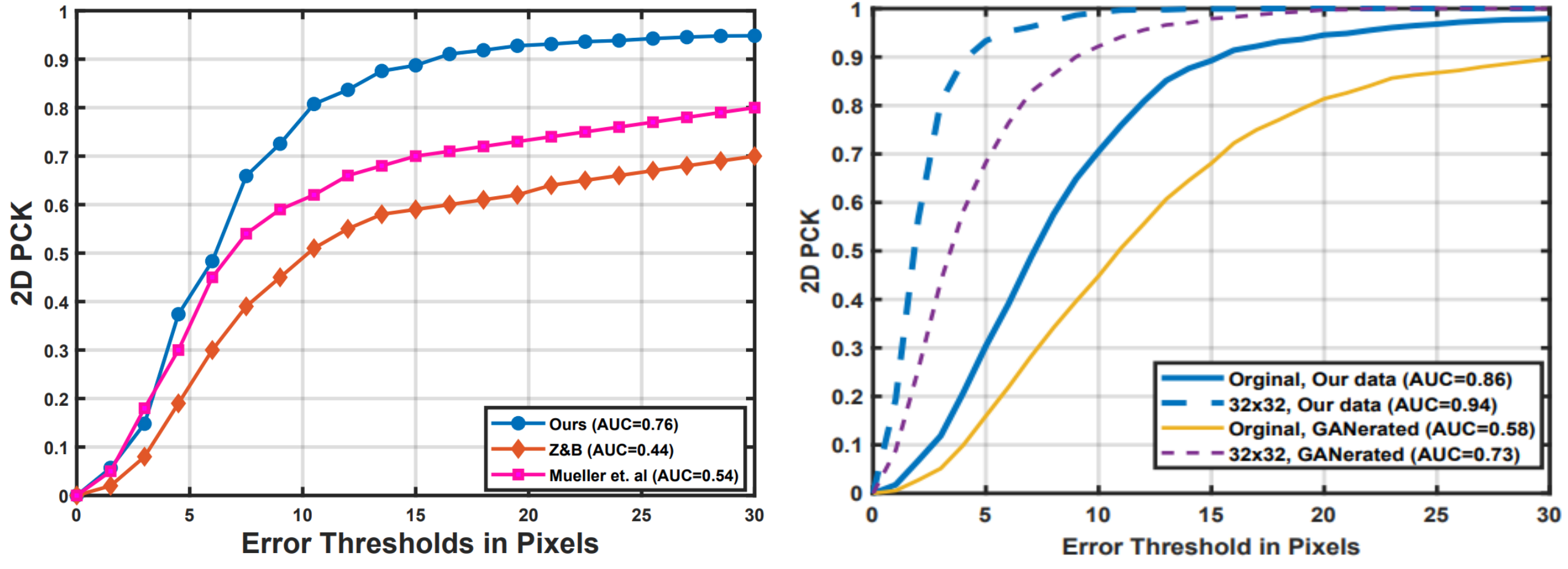}
  \caption{The quantitative results on EgoDexter \cite{mueller2018ganerated} and our \textbf{Ego-Siam} test set.}
  \label{fig:auc_curve}
\end{figure}

\begin{figure*}[t]
  \includegraphics[width=\linewidth]{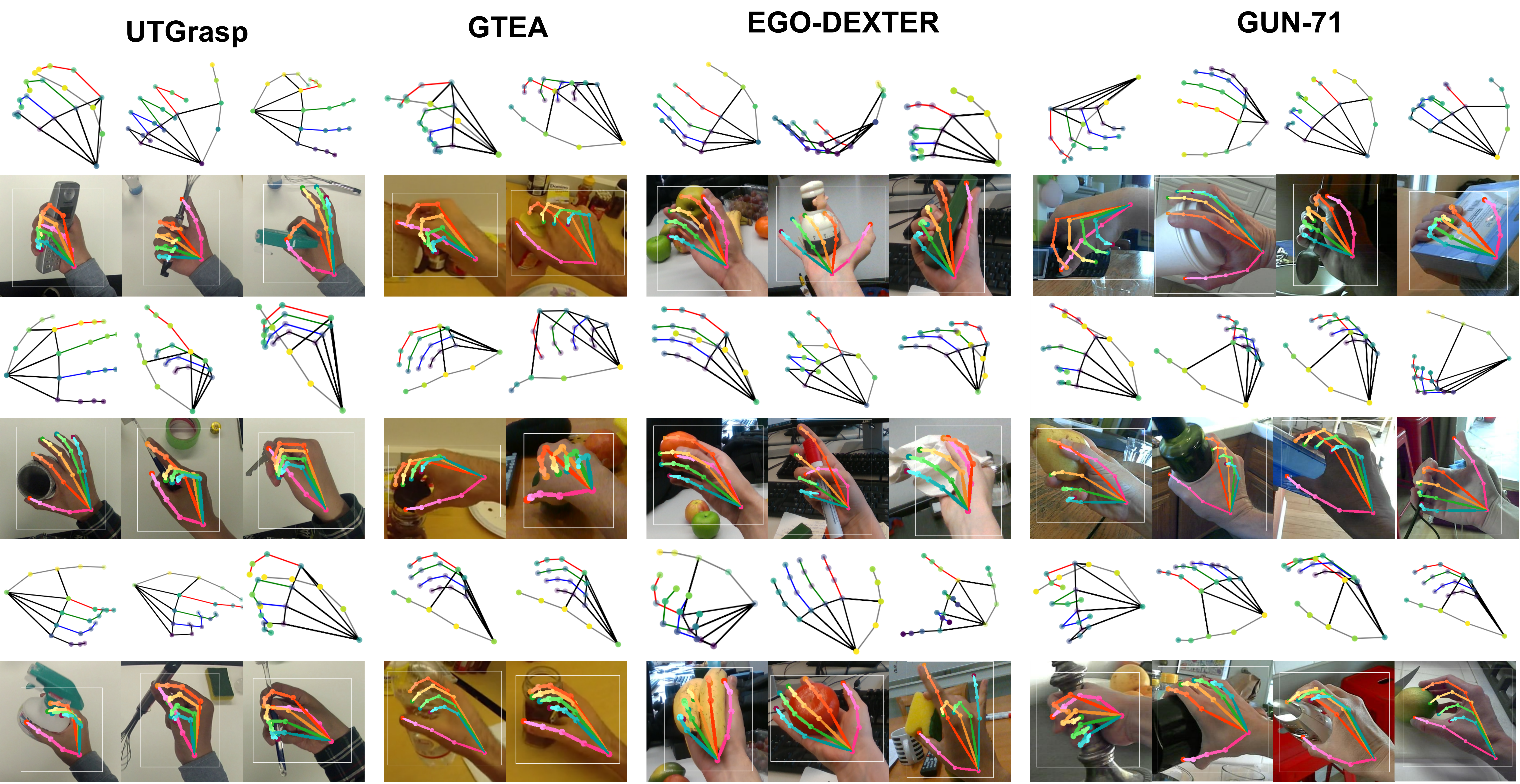}
  \caption{The visual results on UTgrasp \cite{cai2015scalable}, GTEA \cite{fathi2011learning}, EgoDexter \cite{mueller2017real} and GUN-71 \cite{rogez2015understanding} of our 2d/3d hand pose estimator. Our method can work with different hands, light conditions and objects.}
  \label{fig:visual_rst}
\end{figure*}

\subsection{Hand Pose Estimation}
The purpose of this independent evaluation of hand pose module is to highlight the role in hand-object interaction detection system. Same as the evaluation protocol used in all pose estimation works, the Percentage of Correct Keypoints (PCK) score is used as our evaluation metric. It measures whether the given keypoints falls within a pre-defined range around the ground truth. Because our 3-d hand pose is estimated according to the predicted 2-d heatmaps only, we mainly report the PCK score plot of 2-d results. And due to the limited availability of the real hand pose datasets in an egocentric perspective. We can only conduct the quantitative evaluation on EgoDexter \cite{mueller2017real} and our new collected \textbf{Ego-Siam} test set.

The figure \ref{fig:auc_curve} shows the AUC (area under the curve) values on EgoDexter\cite{mueller2017real} and our \textbf{Ego-Siam} data with different setups. Specifically, we compare the 2-d PCK on EgoDexter between Z\&B \cite{zimmermann2017learning}, Mueller et al. \cite{mueller2018ganerated} and ours. Our method greatly outperforms the fingertip detection on EgoDexter dataset. To further evaluate the performance of full hand pose estimation, we also compare the results on our \textbf{Ego-Siam} test set with different training dataset (Ego-Siam training and GANerated \cite{mueller2018ganerated}). The AUC value of training on \textbf{Ego-Siam} outperforms training on GANerated \cite{mueller2018ganerated} on both original image resolution and $32\times32$ (heatmap size) resolution. As for 3-d results, our 3-d hand pose estimation is accomplished by feeding the 2-d heatmap to a simple neural net, we still achieve average error of $\textbf{36.02}$ $\textbf{mm}$ on EgoDexter\cite{mueller2017real} which is a comparative result with their benchmark $\textbf{32.6}$ $\textbf{mm}$. Furthermore, the amount of data we used for training (6k) is much less than Mueller et al. \cite{mueller2017real} (63k) and \cite{mueller2018ganerated} (330k). The results illustrate that our dataset is refined and can be generalised across different datasets. We show the visual results in figure \ref{fig:visual_rst}. It includes the results on UTgrasp \cite{cai2015scalable}, GTEA \cite{fathi2011learning}, EgoDexter \cite{mueller2017real} and GUN-71 \cite{rogez2015understanding}. What surprised us was the performance of our network at GUN-71 \cite{rogez2015understanding}. This dataset contains a large number of different daily activities of egocentric hand-object interactions. It further proves that our method (training with \textbf{Ego-Siam} and our network only) can work across different backgrounds, hands and objects from different datasets.

\begin{figure*}[t]
 \centering 
 \includegraphics[width=1.9\columnwidth]{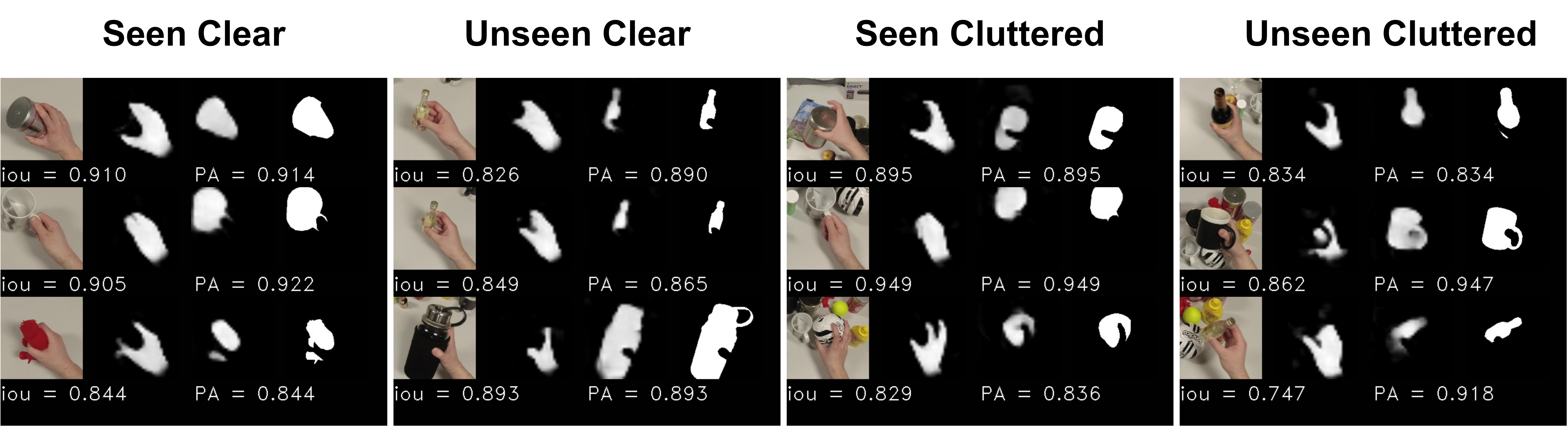}
 \caption{Example of testing results on different conditions. In each set, from left to right are input hand crop, hand segmentation result, object segmentation result and ground truth object mask. 'seen'/'unseen': whether the object is included in the dataset. 'clear/cluttered': whether the background is clear or not. 'IOU': intersect over union between predicted object mask and ground truth object mask. 'PA': pixel accuracy.}
 \label{fig:segmentation_rst}
\end{figure*}

\begin{figure}[t]
\centering
\includegraphics[width=0.9\columnwidth]{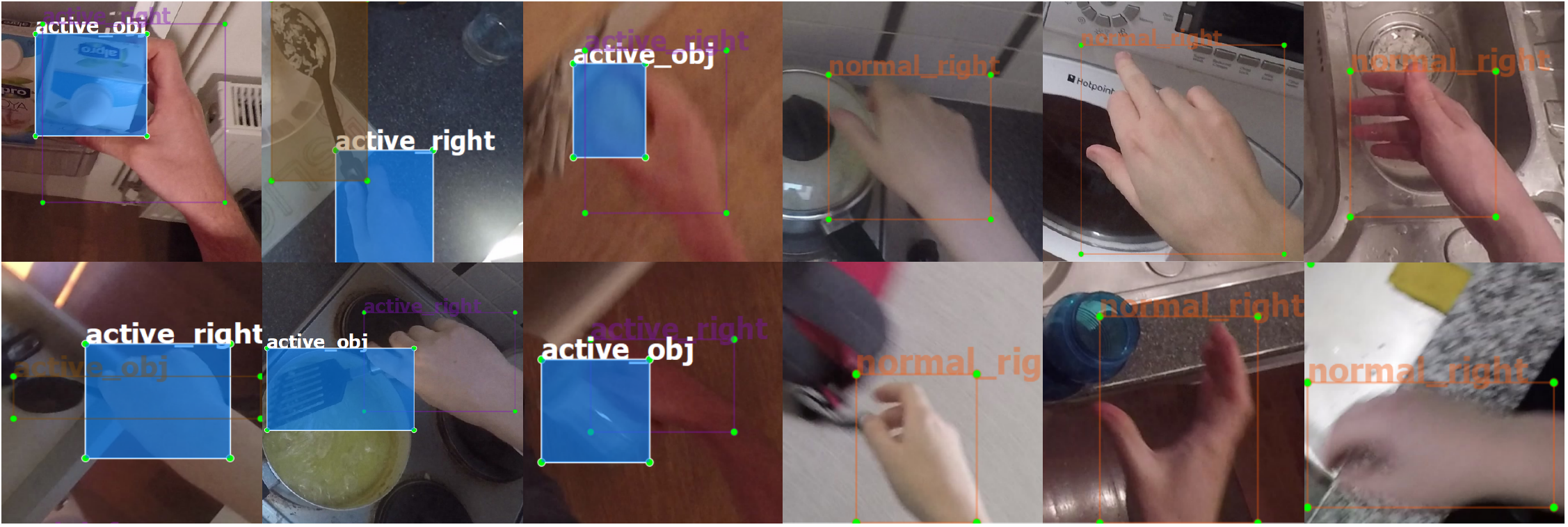} 
\caption{Examples of testing data we re-labelled from EPIC-KITCHENS\cite{damen2018scaling}.}
\label{fig:test_data}
\end{figure}

\subsection{In-Hand Object Segmentation}
To evaluate the performance of our in-hand object segmentation, we manually labelled a test set which contains $400$ images. These images are collected on a table with varies objects placed together. We have intentionally collected data under different conditions which can be categorised as 'seen clear', 'unseen clear', 'seen cluttered' and 'unseen cluttered'. Where 'seen'/'unseen' indicates whether the object is present in the dataset and 'clear/cluttered' means whether the background is clear (contains one object a time) or cluttered (the target object is placed among many objects). We train our network on our GraspSeg (detailed in section \ref{sec:data_anno}) dataset and fine-tune on the newly labelled $3k$ images (detailed in section \ref{sec:data_anno}). The quantitative results over the $400$ test images are shown in table \ref{tab:rst} and the visual results are shown in figure \ref{fig:segmentation_rst}. We use IOU (intersect over union) and PA (pixel accuracy) as the metric of experiments. The quantitative results indicates that both cluttered background and new object (unseen groups) may decrease the accuracy of the segmentation. From the figure \ref{fig:segmentation_rst}, we found sometimes the segmented in-hand object masks are not complete, this could be caused by conspicuous texture change of the objects and these cases usually occurred when dealing with 'unseen objects'. For most cases, however, our algorithm can segment the object in hand quite effectively.

\subsection{HOI Detection Evaluation} 
In this section we focus on evaluating the accuracy of HOI (Hand-object interaction) detection. First, the most straightforward method is to estimate the frame-wise accuracy directly. After which we also further illustrate the accuracy in application scenarios of HOI detection through the first-person action segmentation.

\subsubsection{Test Set for HOI Detection}
As far as we know, there is no widely accepted dataset for egocentric hand-object interaction detection. To facilitate the evaluation of HOI detection from egocentric view, we create a test set by selecting images from the EPIC-KITCHENS \cite{damen2018scaling} dataset. The EPIC-KITCHENS is a large-scale video based dataset contains various of kitchen activities from different people and palaces. All the activities are unscripted and naturally reflect how people interact with objects in daily life. We randomly select about $3k$ frames from the images officially extracted from the videos of $32$ people, and the frames they provided have been auto labelled by the detector from the work of Shan \cite{Shan20}. Because the provided images were evenly sampled from the original videos. Our random chosen images may contain 'no hand' or extremely blurred cases. We manually remove unqualified ones and correct the auto labelled bounding boxes when errors found. The examples of the images are demonstrated in figure \ref{fig:test_data}. We demonstrate the positive samples (with HOI detection) on the left and negative samples (without HOI detection) on the right.


\begin{figure*}
\centering
\includegraphics[width=1.7\columnwidth]{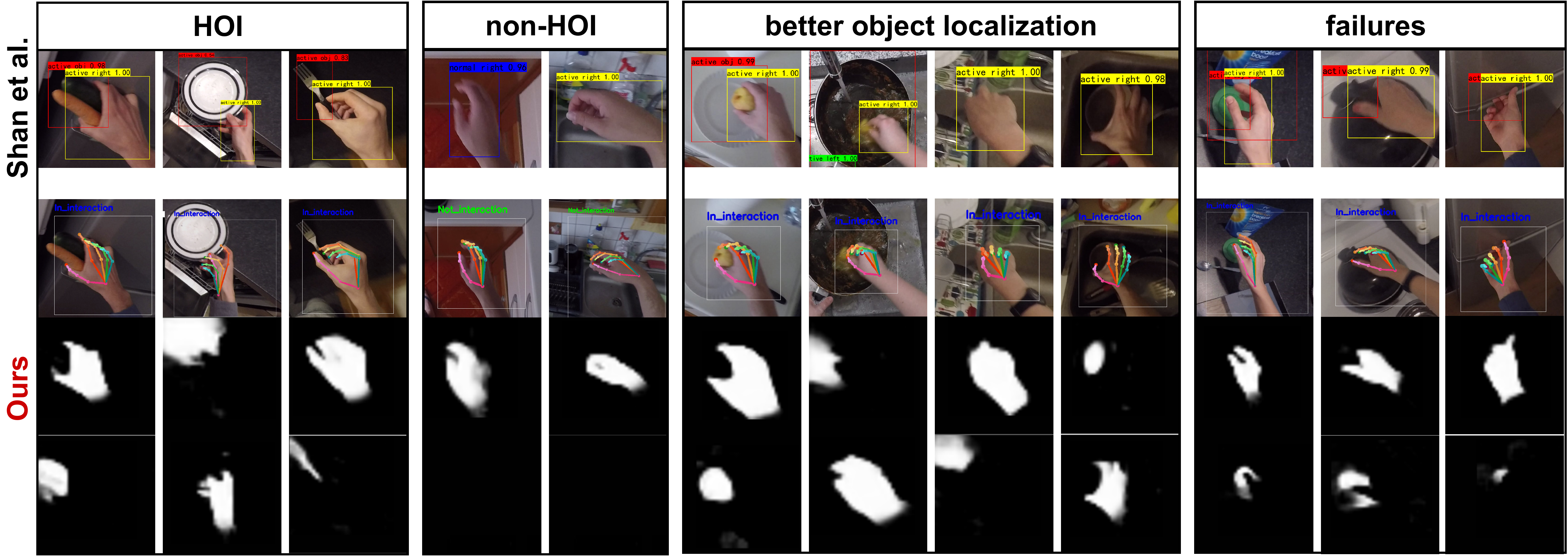} 
\caption{Visual comparison between our method and Shan's \cite{Shan20}. Our method tends to be good at locating HOI with a relatively small object. While Shan's model has better performance on HOI detection with a larger object.}
\label{fig:visual_results_hoi}
\end{figure*}

\begin{table*}[]
\begin{center}
\begin{tabular}{|l|c|c|c|c|c|c|c}
\hline
  & Shan et al. & Ours& No hand pose & No hand mask & No object mask & VGG based \\
\hline\hline
Acc & 0.92 & 0.89& 0.79 & 0.83 & 0.61 & 0.68 (0.66) \\
\hline
\end{tabular}
\end{center}
\caption{The table shows the quantitative results on the test set.}
\label{tab:rst1}
\end{table*}

\subsubsection{HOI Detection on Selected Frames}
\label{sec:HOI_DETECTION}
Specifically, in this experiment, we mainly evaluate:
\begin{itemize}
    \item Our proposed method. We report the results of our proposed method on the test set images.
    \item Shan's model. We report the results detected by Shan's model \cite{Shan20}.
    \item The importance of hand pose. We train our network without the cue of hand pose heatmaps and report the performance of the trained detector. 
    \item The importance of in-hand object mask.
    \item The importance of hand mask.
    \item VGG based image classification. We use a pre-trained VGG as backbone to classify the hand-object interaction status.
\end{itemize}

The results of HOI detection is shown in table \ref{tab:rst1}. It presents the accuracy of different experiment setups. Except the Shan's \cite{Shan20} model, the rest setup are trained on the $6$k images (annotated with green background) and GraspSeg (our newly annotated dataset). The results shows that except the 'no object mask' set, the worst performance comes from the VGG-backbone \cite{simonyan2014very} classifier. The VGG backbone was pre-trained on ImageNet \cite{deng2009imagenet}. It achieves $68\%$ accuracy on HOI detection when only the head parameters were trained, and the performance drops to $66\%$ when all the parameters were involved in training.
The four sets of experiments in the table (column 2-4) illustrate the importance of each of the cues we use in determining hand-object interaction status, especially the cue 'in-hand object mask', without the help of which accuracy drops by $28\%$ directly. Our method achieves $\textbf{0.89}$ accuracy which is comparable to the results from Shan's \cite{Shan20} model. Nevertheless, the size of our model is less than $\textbf{100M}$ (hand pose detector: $\textbf{84}$, and HOI detector: $\textbf{14}$) which is much small than Shan's ($\textbf{361M}$). As for real-time performance, on our old machine (Nvidia Quadro M2000 GPU (4GB) and i5 processor), the Shan's only has $\textbf{1}\sim\textbf{2}$ FPS, while, our model has over $\textbf{30}$ FPS performance.


The visual results are shown in figure \ref{fig:visual_results_hoi}. Both our method and Shan's \cite{Shan20} model perform well on HOI detection. The top row shows the result from Shan's and the above rows are from ours' showing the hand pose detection, hand segmentation and in-hand object segmentation. Our method has better performance on locating the small object being manipulated. Because our method is performed with the cropped hand area, locating the object partially visible in the crop is an apparent shortage.

\begin{figure*}[t]
\centering
\includegraphics[width=1.9\columnwidth]{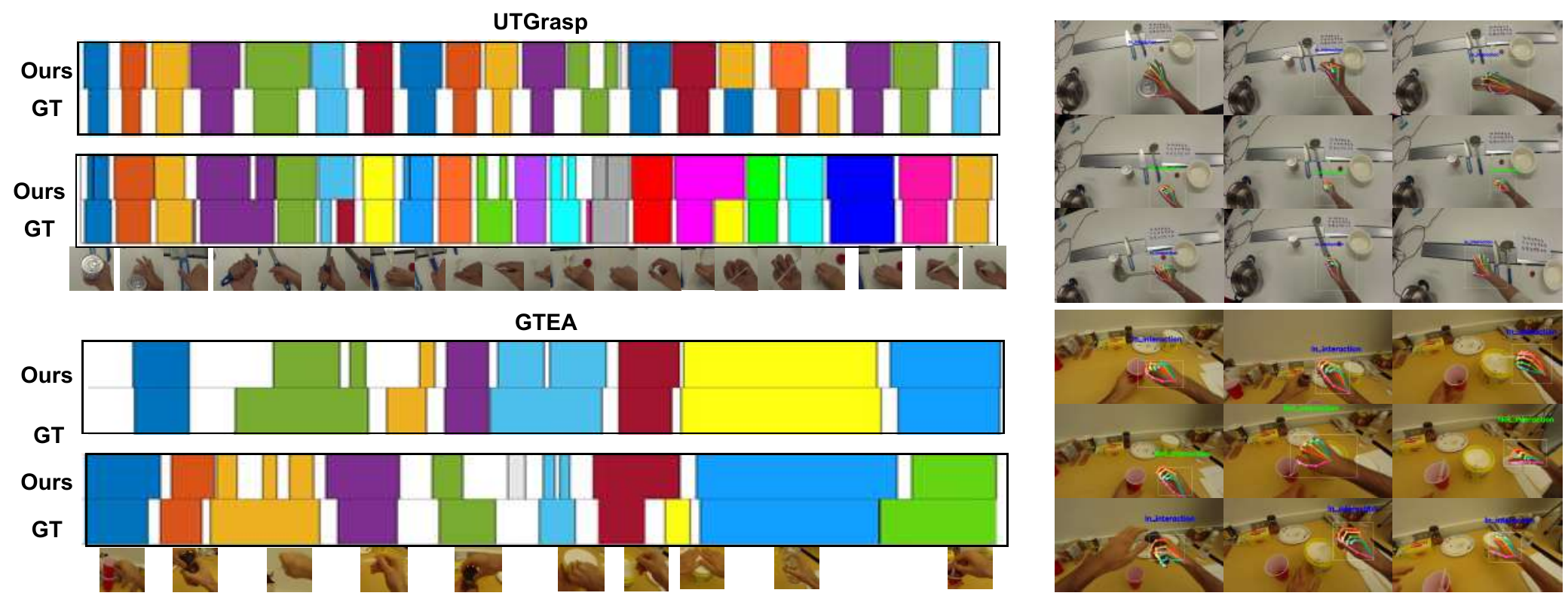} 
\caption{Examples of video segmentation by our HOI system. The left part shows the two example segmentation from each dataset. The right part shows some visual results of our HOI detector.}
\label{fig:seg_rst}
\end{figure*}

\subsection{HOI Video Extraction}
To better demonstrate the practical applications of our detectors. We build a system that can automatically extract the HOI segments from the egocentric video. It should be noted that our detector does not cope with frequent environmental changes and ego-motion. For example, the egocentric dataset like EPIC-KITCHENS \cite{damen2018scaling} contains a large amount of HOI annotations without a clear hand in the scene, which make evaluation pointless. On the contrary, like he prescript dataset GTEA \cite{fathi2011learning} and UTGrasp \cite{cai2015scalable} are performed in front of a table without large-scale ego-motion. This is closer to the ideal condition of delivering guidance with AR/XR technology or analysing the egocentric equipment usage, which are proper for performing the HOI video segmentation.

\subsubsection{Hand Interaction System}
To get our HOI detector running smoothly on video data, we implement a system that contains hand localisation, hand pose tracking and hand-object interaction (HOI) detection. (shown in figure \ref{fig:HOI_system}) For simplicity, the system is designed for working with a single right hand. To \textbf{localise the hand}, we resize the whole frame to $48\times28$ and regard the hand area as a key point. We train a fully convolutional network that outputs a heatmap showing the possible location of hands and their ID (left or right). The ground truth is obtained by putting a Gaussian distribution in the centre of bounding boxes regardless of the original box size. The process can be interpreted as the ROI (region of interest) extraction with a heatmap. \textbf{Hand ID classification} is designed with $3$ kinds of outputs: 'left hand', 'right hand' and 'two hands'. When 'two hands' are detected, we simply take the one on the right side as the right hand. A crop around the detected 'keypoint' is resized and sent to estimate the hand pose, and the \textbf{tracking} is done by re-centring the tracking box to the centre of hand joints over the successive frames.

\subsubsection{Experiments and Results}
For the experimental setup, as the HOI detection system we built solves the problem of hand detection and tracking, for the video data we can then process it frame by frame and record the HOI status. For the GTEA \cite{fathi2011learning} dataset, two hands are involved in most of the interactions. Because the camera wearers are all right-handed, we take the HOI status from right hand only. Both GTEA \cite{fathi2011learning}, and UTGrasp \cite{cai2015scalable} provides the time stamps for actions or object interactions. To facilitate evaluation, we report frame accuracy and F1 score of IOU at $50\%$, which is the same as the evaluation protocol used in Farha et al. \cite{farha2019ms}. We compare the segmentation results with a video-based method MS-TCN \cite{farha2019ms} and Shan's \cite{Shan20} frame-based model. To overcome the noise (false detection) problem of Shan's and our frame-based method, we apply a temporal filter with half-second to smooth the detection results over time. Figure \ref{fig:seg_rst} shows the example results of our HOI system.   We report our results in table \ref{tab:seg_rst}. For UTGrasp dataset, we achieve $89.3\%$ F1 scores on video action segmentation and $82.8\%$ on HOI status accuracy. Our system outperforms the Shan et al. \cite{Shan20} on the action segmentation and has a comparative result on frame-wise HOI status accuracy. For GTEA \cite{fathi2011learning} dataset, our result is very close to the MS-TCN's \cite{farha2019ms}. MS-TCN is a SOTA method of action segmentation that uses the whole video feature to predict the segments.

\begin{table}
\centering
\caption{The $F1$ score with overlapping thresholds (IOU) $50\%$ and frmae accuracy on GTEA \cite{fathi2011learning} dataset. MS-TCN: Multi-stage temporal convolution network. \cite{farha2019ms}. Shan's: Results obtained by running the model provided by Shan et al. \cite{Shan20}.}
\begin{tabular}{ |l|c|c| } 

\hline
     Dataset &UTGrasp & GTEA  \\
\hline
     Ours F1@$50\%$&\textbf{ 89.3}\% & 68.2\% \\
\hline
      MS-TCN  F1@$50\%$&-&\textbf{ 69.8}\%\\ 
\hline
     Shan's F1@$50\%$& 82.1\% & 62.4\%\\ 
\hline
\hline

     Ours Acc& 82.8\% & 71.5\% \\
\hline
      MS-TCN Acc& - &\textbf{76.3}\%\\ 
\hline
     Shan's Acc& \textbf{83.7}\% & 65.6\%\\ 
\hline

\end{tabular}
\label{tab:seg_rst}
\end{table}

\section{Conclusion}
We have presented a work that can detect hand-object interactions from an egocentric perspective. To achieve the goal, we implemented sub-tasks, including hand pose estimation and hand-object pair segmentation. We use a novel data labelling approach for hand pose estimation to tackle the in-hand object occlusion problem and achieve good performance with training on a small amount of data. Also, we annotated a dataset GraspSeg and trained with our novel network for hand and in-hand object mask prediction. We predict the hand-object interaction status with detected hand pose and masks. Our method achieves the accuracy of $89\%$ which is comparable with the SOTA, at the same time, to run in real time. For egocentric video segmentation, we achieves $F1$ score of $0.893$ on IOU at $50\%$ threshold. Our approach shows the great potential on AR guidance and first-person video analysis.

\bibliographystyle{IEEEbib}
\bibliography{example}

\end{document}